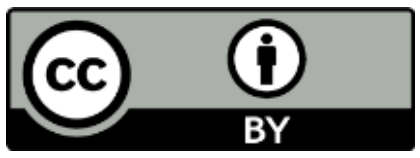

# The methodology of Constructing the Large-Scale Dataset for Detecting Presuicidal and Anti-Suicidal Signals in Social Media Texts in Russian

[1] *I.O. Buyanov, ORCID: 0009-0000-6994-151X <buyanov.igor.o@yandex.ru>*
[2] *D.V. Yaskova, ORCID: 0009-0008-2987-0567 <dary95lobanova@gmail.com>*
[1] *D.S. Serenko, ORCID: 0009-0003-6676-7255 <serenko.d.s@yandex.ru>*
[1] *D.N. Shkereda, ORCID: 0009-0001-5709-1199 <dshkereda@mail.ru>*
[3] *A.D. Yaskov, ORCID: 0009-0004-1952-5445 <ayaskov93@gmail.com>*
[1,4,5] *I.V. Sochenkov, ORCID: 0000-0003-3113-3765 <sochenkov@isa.ru>*

[1] *Federal Research Center "Computer Science and Control" of the Russian Academy of Sciences, 44, build. 2, Vavilova St, Moscow, 119333, Russia.*
[2] *MTS AI, 23, build 5, Podsosenskiy lane, Moscow, 105062, Russia.*
[3] *Yandex, 16, Lev Tolstoy St, Moscow, 119021, Russia.*
[4] *Kharkevich Institute for Information Transmission Problems of the Russian Academy of Sciences, 19, build 1, Bolshoy Karetny Lane, Moscow, 127051, Russia.*
[5] *Ivannikov Institute for System Programming of the Russian Academy of Sciences, 25, Alexander Solzhenitsyn St, Moscow, 109004, Russia.*

**Abstract.** The suicide is a terrifying act of a person who is misled by his own mental state. This problem arises across many countries. Sadly, Russia also has quite high number of persons who committed suicide. Luckily, a subset of these people writes their struggles in social media, allowing a way to find them and help. However, these valuable texts disappearing in many irrelevant texts which is considerably slowing down the decision process about person's suicidal risk. To tackle this problem, in this work we have presented a detailed methodology of building the dataset for detecting texts that describe presuicidal and anti-suicidal signals. This methodology describes the process of instruction and class table creation, the process of annotation, verification and post-annotation correction. Guiding by this methodology, we collect and annotate a large-scale Russian dataset with more than 50 thousand texts from social media. We provide a count statistic of the dataset as well as common problems in annotation. We also conduct basic experiments of building the classification models to show the on go performance on different levels of annotation. Furthermore, we make the dataset, code and all materials publicly available.

**For citation:** Buyanov I.O., Yaskova D.V., Serenko D.S., Shkereda D.N., Yaskov A.D., Sochenkov I.V. The methodology of constructing the large-scale dataset for detecting presuicidal and anti-suicidal signals in social media texts in Russian. Trudy ISP RAN/Proc. ISP RAS, vol. 37, issue 6, part 2, 2025, pp. 191-210. DOI: 10.15514/ISPRAS-2025-37(6)-29.

**Acknowledgements.** This study was conducted in Laboratory “Technologies for analysis and controllable text generation”, additional agreement No. 075-03-2024-490/2 with the support of the Foundation for Assistance to Small Innovative Enterprises within the grant under contract 52ГУКодИИС13-D7/94524. We also thank the reviewers for their valuable comments and all our annotators: Nashuliian Janna, Tiukaeva Anastasiia, Artem Zagidulin, Tatiana Soloshenko, Irina Hmeleva, Leonid Fomin, Alina Riabusheva, Natalia Soloshenko, Denis Martynov, Natalia Matveeva.

# Методология создания большого русскоязычного набора данных для обнаружения пресуицидальных и антисуицидальных сигналов в текстах социальных сетей

[1] *И.О. Буянов, ORCID: 0009-0000-6994-151X <buyanov.igor.o@yandex.ru>*
[2] *Д.В. Яськова, ORCID: 0009-0008-2987-0567 <dary95lobanova@gmail.com>*
[1] *Д.С. Серенко, ORCID: 0009-0003-6676-7255 <serenko.d.s@yandex.ru>*
[1] *Д.Н. Шкереда, ORCID: 0009-0001-5709-1199 <dshkereda@mail.ru>*
[3] *А.Д. Яськов, ORCID: 0009-0004-1952-5445 <ayaskov93@gmail.com>*
[1,4,5] *И.В. Соченков, ORCID: 0000-0003-3113-3765 <sochenkov@isa.ru>*

[1] *Федеральный исследовательский институт «Информатика и Управление» РАН, Россия, 119333, г. Москва, ул. Вавилова, д. 44, стр. 2.*

[2] *МТС ИИ, Россия, 105062, г. Москва, Подсосенский пер., д. 23, стр. 5.*

[3] *Яндекс, Россия, 119021, г. Москва, ул. Льва Толстого, д. 16.*

[4] *Институт проблем передачи информации им. А.А. Харкевича РАН, Россия, 127051, г. Москва, Большой Каретный пер., д. 19 стр. 1.*

[5] *Институт системного программирования им. В.П. Иванникова РАН, Россия, 109004, г. Москва, ул. Александра Солженицына, д. 25.*

**Аннотация.** Самоубийство – это ужасающий поступок человека, которого вводит в заблуждение его собственное психическое состояние. Эта проблема актуальна для многих странах и в России в том числе. К счастью, некоторые из этих людей пишут о своих проблемах в социальных сетях, что позволяет найти их и помочь справиться с их проблемами. Однако эти значимые тексты теряются среди большего количества нерелевантных текстов, что значительно замедляет процесс принятия решения о суицидальном риске человека. Чтобы помочь справиться с этой проблемой, в этой работе представлена подробная методология создания набора данных для обнаружения текстов, содержащих пресуицидальные и антисуицидальные сигналы. Эта методология описывает процесс создания инструкций и таблиц классов, процесс аннотирования, проверки и исправления после аннотирования. Руководствуясь этой методологией, был собран и размечен большой русскоязычный набор данных, содержащий более 50 тысяч текстов из социальных сетей. В работе предоставлена статистика количества данных в наборе данных, а также общие проблемы с разметкой, которые возникли в процессе. Показаны результаты базовых экспериментов по построению классификационных моделей, чтобы продемонстрировать работоспособность на разных уровнях аннотации. Кроме того, набор данных, код и все материалы были сделаны общедоступными.

**Ключевые слова:** создание набора данных; суицид; методология; разметка.

**Для цитирования:** Буянов И.О., Яськова Д.В., Серенко Д.С., Шкереда Д.Н., Яськов А.Д., Соченков И.В. Методология создания большого русскоязычного набора данных для обнаружения пресуицидальных и антисуицидальных сигналов в текстах социальных сетей. Труды ИСП РАН, том 37, вып. 6, часть 2, 2025 г., стр. 191–210 (на английском языке). DOI: 10.15514/ISPRAS–2025–37(6)–29.

**Благодарности.** Исследование выполнено при поддержке Фонда содействия развитию малых форм предприятий в научно-технической сфере в рамках гранта по договору 52ГУКодИИС13-D7/94524, а также при поддержке Министерства науки и высшего образования РФ в рамках государственного задания на оказание государственных услуг в соответствии с дополнительным соглашением № 075-03-2024-490/2 (Молодежная лаборатория “Технологии анализа и контролируемого синтеза текстов”, НИОКТР № 124042600053-7). Мы также выражаем благодарности нашим разметчикам: Насхулиян Жанна, Тюкаева Анастасия, Артем Загидулин, Татьяна Солошенко, Ирина Хмелева, Леонид Фомин, Алина Рябушева, Наталья Солошенко, Денис Мартынов, Наталья Матвеева. Также выражаем благодарность рецензентам за их ценные комментарии.

## 1. Introduction

Although technological advancements have greatly enhanced living standards for numerous individuals across the globe [1]. According to the Federal State Statistics Service [2], in 2019 17 thousand people committed suicide, which is comparable to half of an average Russian regional city. Given findings from recent research indicating rising levels of depression [3], the issue of suicide appears poised to worsen since depression is recognized as a significant contributing factor [4-5]. Beyond causing substantial economic loss for governments, suicide inflicts profound emotional pain upon those close to the victim. The studies show the rise of suicide chance in the social cycle of the suicide victim.

One of the surprising places where suicides can be found is social media platforms such as X, VK, Instagram, and Telegram. On these sites, some people share personal reflections alongside humorous content, occasionally disclosing deeply intimate struggles. For some contemplating suicide, voicing their emotions online acts as a release mechanism alleviating internal pressure. Occasionally, individuals intent on ending their lives leave explicit messages detailing plans, including location and method. Prompt detection of such posts might enable rescue efforts. Even without drastic revelations, tracking subtle signs like emerging depression symptoms or instances of self-inflicted harm could aid intervention prior to severe psychological deterioration.

Over the past few years, there has been a surge in academic publications exploring methods for recognizing various mental illnesses, such as depression or PTSD, using social media data. Methods dedicated to suicide are designed to predict specific outcomes, such as whether someone will attempt suicide during a defined time interval. Unfortunately, there is no way to understand the reasons behind the decision. Also, much of these works are in English. To address this gap, this paper introduces a Russian-language dataset derived from various social media platforms specifically designed to analyze indicators signaling potential suicidal tendencies and suicide chance itself.

To summarize, in our work we present a detailed methodology of creating the dataset that is dedicated to the classification of text that describes factors leading to the higher or lower level of suicide chance. We present the large-scale Russian dataset collected by this methodology that includes more than 50 thousand examples. We also prove the results of building the classification models on this dataset. We made the dataset and models publicly available [6].

## 2. Related Work

The utilization of Natural Language Processing techniques in the field of mental health is made feasible through access to extensive social media data. Commonly employed platforms include Reddit and X, which serve as a source of textual content posted by users. Reddit features specialized sections focused on mental health issues, enabling users to openly disclose their diagnoses, thereby facilitating the creation of high-quality datasets. Similarly, X provides opportunities for users seeking emotional support to publicly share their medical diagnoses. Nevertheless, these posts must undergo verification processes to ensure they do not contain humor, sarcasm, or irrelevant material [7-8].

By employing this strategy, researchers have successfully constructed datasets aimed at identifying users suffering from depression or Post Traumatic Stress Disorder (PTSD) [7, 9], developed datasets highlighting signs of depression [10] leading to tasks such as Early Depression Detection (eRisk), and curated a distinctive suicidal dataset compiled from deceased and surviving X accounts [11]. An exhaustive compilation of existing datasets relevant to mental health studies is detailed in [12].

An alternative approach involves designing questionnaires integrated into prominent social networking platforms. Participants consent to sharing their publicly accessible data, encompassing elements such as status updates, demographic details, and other pertinent attributes. Such methodologies have proven effective in investigating linguistic variations associated with different personality traits [13].

With the rise of large language models (LLM), the researchers start to build the benchmarks in the domain of psychology and mental illness in order to assess the capability of the LLM to tackle these problems. The notable benchmark is PsyEval [14]. This benchmark is a composition of existing datasets providing three assessed dimensions: emotional support skills, diagnosis skills, and knowledge.

Turning attention towards contributions rooted in the Russian language, two valuable works should be mentioned. One study gathers depression-related posts from the platform VK by leveraging a predefined lexicon of depression-specific terms, subsequently analyzing the resultant dataset [15]. Additionally, another investigation compiles essays authored by individuals formally diagnosed with depression, focusing on neutral topics. Through comparative analyses, this study illuminates differences in sets of depression markers observed across depressive narratives versus control samples [16]. Similar work was presented in [17], but for persons who committed suicide and couldn't survive. The work discusses various speech features that characterize the text of these persons. Also, there is a work [18] where the researchers also build the dataset for presuicidal signals but without any granularity.

## *3. The methodology of dataset creation*

The primary task of the dataset is to classify what is being talked about in text into factors that push to or pull from the suicide. We call them presuicidal and anti-suicidal signals, respectively. The idea behind the task is to provide the human the complete and relevant information from the whole text history of the social network account in order to make the decision about suicide status.

Next, we provide a detailed methodology of how we build the dataset. We will start from the description of the data source; next we talk about instruction creation, and finally we outline the general annotation process along with the verification procedure.

### 3.1 The data sources

We can divide data sources into three categories:

- The social networks – we directly parsed the accounts from popular social networks like VK [19] and X [20]. We used custom parsers that use either the API or direct parse of the page. The parsers available on the Github [6].
- The suicidal forums [21] – these are platforms that was popular before the rising of social networks. The main characteristic of these platforms is that they are dedicated to one global topic.
- The existed datasets – we reused some available datasets [15] because they closely match our needs.

Speaking of forums, to our surprise, we found ones that dedicated to suicidal topic in some variants: on some forums there is a general discussion of the suicide, other constructed in the way to provide some online psychological help. We collected the messages where users share their feelings and struggles.

### 3.2 Instruction construction

The instruction consists of two parts – the instruction itself, which describes how to perform tasks, and a table with class descriptions. The main part of the instruction is a document divided into several sections: an introduction, a labeling algorithm, labeling principles, and recommendations. The core text of the instruction is based on the guidelines from a previous study [18]. The introduction provides a general overview of what the annotator will be doing and why it is necessary. The latter serves as a strong motivational element, as the annotator is informed that the quality of their work will directly impact future assistance efforts. During discussions with annotators, substantial feedback was collected, with many emphasizing the importance of the project.

Following the introduction is the annotation algorithm, which describes the purely technical actions that annotators must perform to carry out their work. Next is a detailed, step-by-step description of how to begin the annotation process and how the annotation should be performed. The final section is the annotation principles block. This is a set of rules that must be followed when making a decision about selecting a particular class. The second part of the instructions consists of class tables describing all the classes included in the annotation. The first table contains presuicidal classes, while the second contains anti-suicidal classes. The complete table includes:

- The class name, which consists of two parts: the group and the class itself;
- The class description, indicating what exactly should fall under this class;
- Examples containing texts that correspond to the specific class.

It should be noted that the algorithm provides references to two tests: the Emotional Well-Being Test [22] and the Beck Depression Inventory [23]. These tests are intended to monitor the emotional state of annotators, as they have to work with texts of an extremely negative and distressing nature. The assumption was that if test results showed a negative trend, the annotator should stop performing the work.

To create the table of classes, we reviewed literature on suicidology, specifically [24]. The study revealed that there are actually numerous models of suicide. Typically, they describe similar factors and signals (indicators), but differ in how these elements are grouped. Sometimes, weakly distinguishable phenomena were included as separate signals. We decided to generalize several systems and create maximally atomic signals in such a way that, on one hand, they would be easily distinguishable in meaning, and on the other hand, other researchers could adapt them to their theoretical models. A natural limitation was that the signals had to be expressible in social media texts.

We selected the systems from [18], as they are well-formed and represent approaches that vary in time and culture. The work [18] was done by the demands of helpers that do investigate social media seeking persons with a high suicidal chance. In addition, they all are conveniently described in tabular format or as lists. The "mood board" technique was applied to synthesize the classes. All lists and tables were placed in a single space so that they could be viewed at a glance. Then, the following iterative process was carried out:

1. The entire field was reviewed;
2. Based on emerging associations, a feature was formed;
3. Both the feature itself and signals from other systems were marked with color;
4. The list of formed features was checked for consistency, adequacy, and completeness;
5. The process returned to step 1 until the list was fully converged.

The results of anti-suicidal classes are few in number. They were well described in the [25] system and all formed the basis of the table.

The latest version of presuicidal signals includes 33 classes grouped into 7 categories.

With the correction of the anti-suicidal classes, which we will describe later, the final version of anti-suicidal table contains 12 classes without division into groups.

## 3.2 Instruction Validation

After developing the main annotation guidelines and class tables, we conducted several development annotation rounds within our team:

- Annotation of presuicidal signals by two ML specialists on 200 examples.
- Annotation of presuicidal signals by three non-specialists on 200 examples.

- Annotation of anti-suicidal signals by two ML specialists on 100 examples.
- Annotation of anti-suicidal signals by three non-specialists on 100 examples.

The non-specialist means that our team members didn't participate in guideline creation nor see the data. Thus, they simulate new annotators. The difference in sample size was due to the significantly lower number of anti-suicidal classes compared to presuicidal ones. ML specialists and non-specialists annotated the same dataset, respectively. After the annotation by ML specialists, their labels were compared to identify discrepancies. Each such case was analyzed individually, resulting in 28 revisions to the guidelines and presuicidal class table. After implementing all changes, a trial annotation was performed by non-specialists. Follow-up meetings were held with each annotator to discuss difficulties and observations. Based on the feedback, further revisions were made to the guidelines and class tables. Psychological well-being was also assessed—no significant negative effects were observed. The anti-suicidal class annotation followed the same process, requiring fewer than five revisions. Four new anti-suicidal classes were introduced. After finalizing the guidelines and class tables, the main development phase was completed. The guidelines underwent minor adjustments during the annotation process.

During the annotation, we see that a notable part of the examples has multiclass annotation. This brings some special features to how to calculate the inter-annotator agreement and how to aggregate the annotation from multiple annotators.

To automatically calculate inter-annotator agreement, we used Krippendorff's alpha [26]. Since the dataset annotation allows for multiple classes per example, we applied a specialized metric for Krippendorff's alpha called "Measuring Agreement on Set-Valued Items" (MASI) [27]. It is calculated using the following formula:

$$MASI = \frac{n(A \cap B)}{n(A \cup B)} * M$$

Where A and B are label sets, n is a number, and M is a monotonicity scaler, determined by the following rules:

- If the sets are identical, M equals 1.
- If one set is a subset of the other, M equals two-thirds.
- If the intersection of the two sets is non-empty, M equals one-third.
- Otherwise, M equals zero.

For agreement calculation, we used implementations of Krippendorff's alpha and MASI from the NLTK (Natural Language Toolkit) library [28]. The alpha values for abovementioned rounds before any corrections are shown in Table 1.

*Table 1. The annotator agreement in development annotation.*

| Signal type | Annotators | Krippendorf's alpha value |
|---|---|---|
| Presuicidal | ML specialists | 0.39 |
| | Non-specialists | 0.39 |
| Anti-suicidal | ML specialists | 0.45 |
| | Non-specialists | 0.41 |

We tested three aggregation methods on the non-specialists' annotations:

- Full agreement–cases where all three annotators agreed.
- Standard majority voting–a text receives a class label if at least two annotators assigned it. For multi-class cases, the label sets must fully match between at least two annotators.
- Soft majority voting–a text receives labels that appear at least twice in the combined list of annotations. Multi-class labels are split into individual classes for counting.

After each aggregation method was applied, we selected 30 examples randomly and checked for compliance with the class table. Agreement was measured as the percentage of examples where the label was deemed correct by the guideline author. For aggregation methods we also report coverage, the percentage of non-empty class assignments. The results were as follows:

- Non-specialists: 95%, 75%, 55% agreement.
- Full agreement aggregation: 100% agreement, 33% coverage.
- Standard majority voting: 91% agreement, 71% coverage.
- Soft majority voting: 91% agreement, 93% coverage.

We see that agreement between the guideline author and the non-specialist varies notably from a well-performed level to a moderate level. The full agreement, as expected, shows full agreement with low coverage. On the other hand, we see that soft majority voting has higher coverage compared to standard majority voting, keeping the same level of agreement. So, we decided to use this variant of aggregation.

Notably, during post-annotation discussions, annotators provided justifications for their labels in many cases where they disagreed with the guideline author. This highlights that the task has a high level of subjectivity that needs to be accounted for. In Table 2 we present some examples where annotators' decisions completely mismatched.

*Table 2. Examples of subjective texts. Texts are taken from the corpus. We added the comments at the time of writing.*

| Subjective text | The comment |
|---|---|
| У меня трясутся руки, как у последнего алкаша (My hands are shaking like a drunkard's). | Does the user really have a trouble with an alcohol or is it just a speech figure describing the fear or is it perhaps physical health issue? |
| Я был лишь виртуальным для неё (I was just a virtual person to her). | Some of the annotators recognized it was a reflection of relationship problems, some decided it was not significant to be it. |
| Вот бы каждый человек, который разрушил мою менталку оплачивал бы мне сеанс у психолога (I wish every person who ruined my mental state would pay for my therapy sessions). | Again, is it just a speech figure or does this person really have problems with mental health? |

## 3.3 Annotation Team Formation

We used two primary channels to find annotators. These are trusted professional contacts and freelance platforms, aggregator platforms where clients post tasks and freelancers apply.

A total of eight annotators were recruited: three for the test set and five for the training set. All annotators underwent the following onboarding steps:

1. Orientation meeting: Overview of the project, tasks, working conditions, and annotation interface. Annotators could ask questions.
2. Training annotation: 100 pre-annotated examples from the internal team.
3. Feedback session: Discussion of training annotations and clarifications.
4. Rapid verification cycle: During the first iteration, every 500 annotations, an ML specialist reviewed 30 newly labeled examples and provided feedback.
5. Training annotations were manually reviewed due to task-specific complexities.

Performance was evaluated at two levels:

- Correct examples: Cases where soft majority voting with the original annotation yielded a non-empty class.

- Acceptable examples: Correct examples plus annotator labels deemed valid by ML specialists. That is because of task subjectivity.

The average correct example rate was 54.2%, and the acceptable rate was 73.5%. One freelancer, achieving an 80% acceptable rate, was assigned to the test set. During rapid verification, the average error count was 5 after the first check and 2 to 3 thereafter.

## 3.4 Data Sampling Schemes

Our goal was to annotate 50 thousand examples. To improve annotation efficiency and control class distribution, we split this value into 5 iterations by 10 thousand, and further, each iteration was split into 5 blocks (2 thousand examples). One block is one annotation task.

We apply different sampling schemes for each iteration. Sampling was based on pre-extracted features. The sampling includes standard selection criteria and optional ones. The former applied to each iteration in such a way that ensured the uniform distribution of basic statistics. The latter is used to influence the resulting class distribution.

The standard criteria include:

- No foreign-language texts.
- Balanced text length distribution.
- Presence of first-person pronouns.

Optional criteria include:

- Predictions from a suicidal signal detection model [29].
- Subsets of emotion model predictions [30].
- Sentiment model predictions [31].
- Keyword sets.
- Etc.

## 3.5 General Annotation Process

The annotation workflow for a single annotation block consisted of the following steps:

1. Data sampling: Examples are sampled from the corpus using an ML specialist-approved scheme.
2. Data distribution: Samples are uploaded to Label Studio [32] and distributed across annotators.
3. Initial screening: First-time annotators complete the Beck Depression Inventory (BDI).
4. Pre-annotation check: Annotators take an emotional well-being test.
5. Annotation: Annotators label the data.
6. Post-annotation check: Annotators retake the emotional well-being test. If it was their final task, they also complete the BDI.
7. Debriefing: ML specialists hold meetings to discuss questions, suggestions, and emotional venting–a simple psychological support method to alleviate stress.
8. Guideline adjustments: ML specialists revise the guidelines as needed.

## 3.6 Annotation verification

To ensure the quality of annotated data, it was necessary to develop a data verification process. It was decided to conduct verification in parallel with the primary annotation, meaning that instead of checking the data post-annotation, the annotation of the control set was performed by the instruction creators. This approach has the advantage of saving time, as the control set annotation is conducted concurrently with the primary annotation. Given the subjective nature of the annotation, as noted earlier, it is advisable to manually review mismatched examples obtained after comparing the primary annotation and the control set to obtain a more accurate assessment. Thus, the verification algorithm is described as follows:

1. From each annotation task, a subset of data is selected to form a control set.
2. The annotation of the control set is performed by the instruction creator.
3. If necessary, make adjustments to the instructions.
4. Upon completion of the primary annotation, compare the data in the primary and control sets.
5. If the number of mismatches exceeds a predefined threshold, conduct additional verification of discrepancies between the annotator's and verifier's responses.
6. If the number of mismatches still exceeds the threshold, perform an error analysis and return the task to the annotator for revision.
7. If the number of mismatches does not exceed the threshold, the task is accepted.

Based on the review results, a meeting was held with the annotator to discuss the issues. Subsequent checks showed that no further corrective meetings were required. Since the task is subjective and involves a large number of classes, it was decided to evaluate not by specific classes but by signal direction, of which there are three: anti-suicidal, pre-suicidal, and irrelevant. It was decided to adjust the dataset based on model training results, while at this stage ensuring a clear separation into the three main signals. Evaluating at this level allows for sufficiently clear class boundaries, as anti-suicidal and pre-suicidal signals are opposites, and irrelevant signals are also clearly distinguishable simply by contrast with the target signals.

Based on the conditions, experience with data annotation, and other studies [33] involving subjective tasks, the acceptable mismatch threshold for control annotation was set at 15%. The number of examples selected from each block (consisting of 2000 examples) was 185 (9.2%). This "uneven" number was chosen during initial design when the certain degree of subjectivity was not yet known, with provisions for alternative verification methods. A total of 5 verification tasks were annotated for the test set and 25 tasks for the training set. Additionally, 2 tasks were completed to resolve annotations in the test set where soft aggregation failed to assign class labels. The volume of one task was 357 examples. The upper mismatch rate across all control tasks for the training set was 14.55%, and for the test set, it was 13.99%. Thus, both test sets met the established threshold.

## 3.7 The annotation correction process

After completion the annotation process, we started to analyze the whole dataset and experiment with the models. Two main problems were found:

- Based on confusion matrix of presuicidal and anti-suicidal model we figure out that the irrelevant texts heavily intertwine with significant part of the classes.
- The metrics of the anti-suicidal model was significantly lower compared to presuicidal model.

Analysis of models revealed the sources of the first problem:

- The 3rd person rule violation – the texts where the signal is not related to the author should be considered as irrelevant.

- ✓ Example: «Она сказал, что хочет поскорее сдохнуть уже» (She said she wanted to die quickly.)
- The relatively complex texts where the model is triggered by some lexical parts but actually there is no clear definition of the signal.
  - ✓ Example: «И говорить что настоящие страдания только скрытые от всех - это как говорить что те, кто реально хочет умереть уже в гробу» (And to say that real suffering is only hidden from everyone is like saying that those who really want to die are already in their coffins.)
- The texts that have multiple classes.
  - ✓ Examples: «Мне настолько всё это надоело, я не знаю, что мне делать, я постоянно думаю о смерти» (I'm so tired of all this, I don't know what to do, and I keep thinking about death.)
- Errors in the annotation.

To find the examples that are dubious, we applied the dataset cartography technique [35]. On the given map that is presented in Fig. 1, we can estimate the hard-to-learn region as a quadrant x < 0.4 and y < 0.4.

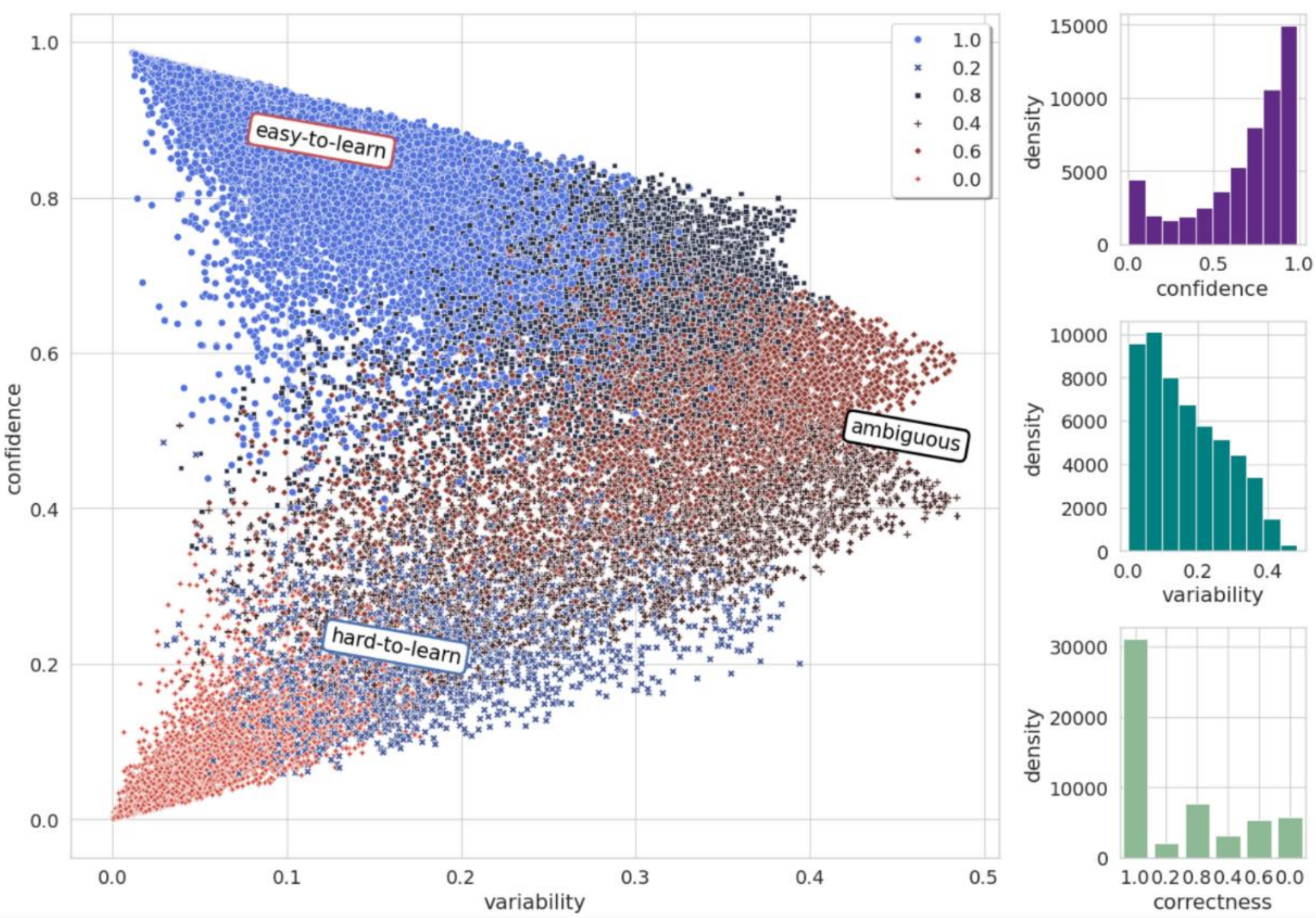

*Fig. 1. Data cartography for the exact presuicidal dataset.*

The count class distribution of selected examples is shown in Table 3. As we can see, the top class is an irrelevant class. Besides anti-suicidal signals that, as we know, have issues, the rest of the classes are from the feeling group. This is not surprising, as the feelings interpretation is heavily subjective by nature. Thus, we decide to reannotate the whole irrelevant examples that don't come from the test set because the analysis shows that these examples are mostly correct but complex.

*Table 3. The class distribution in hard-to-learn region.*

| Class name | Count |
|---|---|
| Irrelevant | 1,726 |
| Anti-suicidal signal | 992 |
| Feelings/mental emptiness, depression, longing, sadness | 860 |
| Feelings/negative self-image, guilt, shame, worthlessness, self-flagellation | 763 |
| Feelings/helplessness, hopelessness, hopelessness, despair | 710 |

The deep analysis of the anti-suicidal model reveals that there is too much noise in the annotation that mainly comes from too abstract names of the classes. We decide to reassemble the class table of anti-suicidal classes by the next procedure:

1. 20 examples per existed class are selected.
2. The main sense of each text is annotated by ML specialist. Each specialist forms his own set of main senses.
3. The result sets are cast to the close set.
4. The two close sets are aligned to each other and grouped together.
5. The final set of groups becomes a new class table.

As an effect, several new classes are emerged saving almost all source classes but with clearer semantic meaning.

Having the new class table, we reannotated the anti-suicidal dataset. Because of the limited budget and time, the annotation was done by different schema. We decided to validate the annotators each 500 annotated examples by the batch of 50 texts. The threshold of the errors was 8 examples (15%). All checked examples become a test part (10%), while rest of the examples become train (80%) and validation parts (10%).

## 4. Dataset statistics

We collected 57,810 annotated examples in total. The presuicidal dataset consists of 38,406 examples, anti-suicidal dataset consists of 9702 examples, irrelevant examples are 9702 (not a mistake). Table 4 and Table 5 show the top 5 counter distribution of presuicidal and anti-suicidal dataset respectively. The final Krippendorf alpha for the test part of presuicidal dataset is 0.542. As we mentioned earlier, the significant part of the dataset is a multilable examples. The ratio of such examples is 0.261.

## 5. Experiments

In this section we present the results of the model learning on the final dataset. As we mentioned earlier, the structure of the classes allows us to learn the model with different levels of granularity. We hypothesize that with a high level of granularity the model will better capture the difference between classes as semantic change will be clearer. Besides the plain granularity, one might explore more complex schema where some classes are collapsed to a group and some classes rest as exact labels. It's useful if one might want to isolate some dubious classes to make the model more robust. An example of such a group is feelings.

We show results for the RuBERT [36] model only, as it constantly shows better performance compared to the RoBERTa [37] and DeBERTa [38] models.

*Table 4. The class distribution of anti-suicidal dataset.*

| Class name | Count |
|---|---|
| Having positive social connections | 1,650 |
| Expression of love | 1,384 |
| Expression of happiness, joy, contentment | 858 |
| Positive self-assessment | 595 |
| Expression of love; Having positive social connections | 486 |

*Table 5. The class distribution of the presuicidal dataset.*

| Class name | Count |
|---|---|
| Death/thoughts about death | 4,205 |
| Feelings/negative self-image, guilt, shame, worthlessness, self-flagellation | 3,236 |
| Problems in the outside world/unhappy love, problems with friends, difficulties in building relationships | 2,602 |
| Feelings/helplessness, hopelessness, hopelessness, despair | 2,359 |
| Feelings/mental emptiness, depression, longing, sadness | 1,964 |

We do basic text processing like text lowering and removing all non-alpha symbols. We also remove all multiclass examples for this experiment. The classes that have less than 100 examples were also removed. We derived the datasets from the whole data that we named as the master dataset. That means that irrelevant examples are shared across the presuicidal and anti-suicidal parts, while they actually are different, as they were sampled by different sampling schema and were annotated separately. So, the results might be biased. Table 6 shows the result performance on the different granularity levels.

As for the presuicidal model we see, that exact granularity is significantly lower than group granularity, and, on the other hand, the ternary model with irrelevant, anti-suicidal, and presuicidal classes with the binary model (relevant and irrelevant) shows even better results. That supports our hypothesis that the higher the level of classes, the better the model can distinguish them. We also see that the group presuicidal model performs on par with the anti-suicidal dataset.

*Table 6. The performance of the RuBERT on different dataset settings.*

| Model name | Label granularity | Class count | Precision | Recall | F1-score |
|---|---|---|---|---|---|
| Presuicidal | Group | 8 | 0.65 | 0.65 | 0.65 |
| | Exact | 26 | 0.61 | 0.51 | 0.53 |
| Anti-suicidal | Exact | 9 | 0.70 | 0.59 | 0.63 |
| All | Binary | 2 | 0.71 | 0.71 | 0.71 |
| | Ternary | 3 | 0.71 | 0.69 | 0.70 |

The analysis of the errors reveals that the main problem is that the irrelevant texts are notably intertwined with several classes. Manual investigation shows the same problems that were highlighted in the annotation correction section above. Especially the problem of overlapping lexical features and the 3rd rule violation.

## 6. Annotation instruction

Here we provide the annotation instruction translated into English along with the lists of presuicidal and anti-suicidal signals. Due to space limitations, we omit the instruction part dedicated to the annotation interface as well as examples for each class. You can find the full version of the instruction and class tables in our repository [6].

## 6.1 Introduction

You will be shown texts collected from various social networks. It is necessary to distribute them according to the classification below. The annotation is needed to create a classifier that will make it easier for volunteers to find people who are on the verge of suicide so that they can be consulted and provided with assistance. Without exaggeration, we can say that by doing this work, you are contributing to saving someone's life.

## 6.2 How to annotate

It is necessary to study the description of pre-suicidal signals – signs in the text that indicate a possible suicide – by clicking on the [link to class table]. The description of anti-suicidal signals by clicking on the [link to class table]. The latter reduces the likelihood of suicide.

Before you start marking, you need to complete two tests.

- The emotional well-being test is completed before and after each marking block.
- The Beck Depression Scale is completed before starting all tasks and at the end of all tasks.

(The instruction of interacting with Label Studio is omitted)

## 6.3 Annotation principles

1. Despite data cleaning, you may encounter texts that violate the laws of the Russian Federation. We are not responsible for such texts, as they were collected from social networks. You may notify us if you come across a text that violates Russian laws via the feedback form (or messenger).
2. The content of the text must relate to the author. The text should contain first-person personal and/or possessive pronouns (I, we, my, our, me, us). For example:
   a. “I want to escape from this oppressive outside world into myself,”
   b. “Well of course I am loved, but what’s the point of being with someone who isn’t.”

   Or other speech patterns that indicate the content refers to the author. For example, “and a month later you write, ‘Kostya, I’m sorry, I met someone else, I truly love him’” – it can be said that the author received such a message. If a person talks about someone else, this information is irrelevant. Examples of irrelevant texts:

   c. “A person needs another person, or a chocolate with tea is fine too” – it cannot be determined that the author needs chocolate and tea.
   d. “Oh, Atsumu really wanted to yell at his brother, to vent his emotions and find out, to ask about all the feelings between them, and not +” – this is about a character.
   e. “Everyone has something that is most important to them.” – this refers to an abstract group of people.
   f. “She’s all dressed up with earrings and eyeshadow FOR BREAKFAST, and her husband of course looks like a water balloon that you don’t throw away for at least three years.” – this refers to third parties.
3. Messages in a foreign language should be marked as unrelated to suicide.
4. You must not attempt to interpret texts in any way, for example, assuming that the person wrote the text to attract attention. You must judge strictly according to the class descriptions. It is also not allowed to assess texts based on perceived context. Example: “but I didn’t know how or what – all my plans for the future were connected with him, in fact everything was connected with him – favorite books, favorite music." Someone might think the author is describing negative relationships, because instinctively, if someone talks about their relationship this way, it probably ended badly. However, the text does not directly state whether the relationship ended, or whether it was bad at all.

5. If a signal is described as having happened in the past, it should also be labeled. However, for classes related to feelings, this principle applies weakly.
6. If you feel emotionally distressed or it becomes difficult for you to continue the task, you should stop immediately.
7. The texts may have unpleasant subjects; try not to annotate them right before sleep.
8. To clarify disease codes in the format F{number}.{number}, use this link to ICD-10: https://mkb-10.com/
9. If there is one clear class and it's unclear whether another class is present (especially common in short texts), do not try to look for it or spend time on it.

## 6.4 Specific points for dataset annotation

1. If there is an example describing a breakup, and it is not clear whether it is about a husband and wife, you should assign the class "External world problems/Unhappy love, problems with friends, difficulties building relationships"
2. If, due to some comparison, allusion, metaphor, etc., you cannot clearly determine the class, you should assign the irrelevant class. Example: "God, I must have been Commander Shepard in a past life, given how lucky I am with people in this one."

## 6.5 List of classes

### 6.5.1 List of presuisidal classes

1. Clinical manifestations/Depression – this class includes texts with facts or mentions of the diagnosis "depression", as well as symptoms of depression.
2. Clinical manifestations/Insomnia – a condition that prevents a person from falling asleep.
3. Clinical manifestations/Eating disorder (ED) – a range of behavioral syndromes associated with disruption of the eating process.
4. Clinical manifestations/Fatigue, stress, resource depletion – manifestations of physical and/or emotional fatigue, stress, and a state of exhaustion of physical and/or emotional resources.
5. Clinical manifestations/Anxiety, fear, phobias, obsessive thoughts – anxiety is a negative emotional state, expectation of trouble, impending danger, or unfavorable events, disproportionate to the actual situation, as well as sensations of fear and worry.
6. Clinical manifestations/Other mental disorders – this class includes texts mentioning any mental disorders not specified in the current list.
7. Clinical manifestations/Physical illnesses, disability – texts mentioning physical illnesses, injuries, or disability fall into this class.
8. Clinical manifestations/Past suicide attempt – this class includes texts mentioning suicide attempts in the past.
9. Clinical manifestations/Crying, hysterics – this class includes texts describing crying, tears, sobbing, hysterics, etc. (do not confuse with tearfulness).
10. Destructive behavior/Self-harm – this class reflects behaviors and thoughts related to self-harm, cutting, or inflicting physical pain on oneself.

11. Destructive behavior/Alcohol and drugs – problems and behaviors associated with alcohol, drugs, or other psychoactive substance abuse.
12. Destructive behavior/Problems with the law, discipline – facts from the past or present related to problems with the law or discipline.
13. Family problems/Bullying, physical abuse – this class includes all types of physical violence within the family and at home.
14. Family problems/Sexualized violence – events in the family environment related to sexualized violence.
15. Family problems/Family breakdown – facts related to the breakdown of the family in any form.
16. Family problems/Difficult relationships with relatives – in this class, we are interested in emotional relationship difficulties (not physical violence).
17. Family problems/Pregnancy difficulties, abortion, miscarriage – this class includes texts mentioning problems related to childbirth or pregnancy.
18. External world problems/Bullying, physical abuse – events at school, among peers, at work, or outside the home, related to bullying, emotional and physical humiliation, or violence.
19. External world problems/Sexualized violence – events at school, among peers, at work, or outside the home, related to sexualized violence.
20. External world problems/Unhappy love, problems with friends, difficulties building relationships – negative emotional states and feelings associated with situations of unhappy or unrequited love, as well as problems in friendships.
21. Death/Death, accident or suicide of loved ones – loss, death, or suicide of family members or close people, including friends.
22. Death/Thoughts about death – permissive attitudes and thoughts about suicidal behavior.
23. Death/Suicidal intentions – messages containing a concrete plan of action. Differs from "desire to die" in the declaration of actions.
24. Feelings/Helplessness, hopelessness, despair – negative emotional states, beliefs, and situations associated with feelings of helplessness and hopelessness.
25. Feelings/Negative self-perception, guilt, shame, worthlessness, self-blame – negative emotional states, beliefs, and situations associated with negative self-perception, feelings of guilt, shame, or worthlessness.
26. Feelings/Loneliness, misunderstanding, isolation, abandonment – negative emotional states, beliefs, and situations associated with feelings of loneliness, sadness, resentment, misunderstanding, etc.
27. Feelings/Aggression, anger, resentment, rage, irritability, jealousy, injustice – expression of aggression, anger, protest, rage, and other active negative emotions.
28. Feelings/Emotional suffering, emptiness, depression, melancholy, sadness – this class includes texts mentioning emotional suffering (without specifics), pain (if not in the context of physical pain), the feeling that "everything is bad", etc.
29. Other/Destructive attitudes – this class includes worldviews such as fearlessness of death, devaluation of life, reflections on the futility of life.
30. Other/Stalking – stalking is when a person is obsessively pursued. An ex-boyfriend or girlfriend, a rejected admirer, or even a stranger may wait at the entrance, watch, ambush at work, write letters, call persistently and systematically on personal or even work phones, send gifts – all these are manifestations of stalking.

31. Other/Problems with career choice and realization – this class includes texts mentioning forced dismissal, unemployment, long unsuccessful job searches, dissatisfaction with current job, or inability to find oneself in a profession.
32. Other/Money problems (poverty, debts) – temporary or permanent situations with debts, poverty, or destitution.
33. Other/Imitating an idol in suicidal behavior – this class includes texts expressing sympathy or imitation of friends, book or film characters with suicidal themes, as well as imitating an idol or public figure in suicide.

### 6.5.2 List of antisuisidal classes

1. Protective factors/Expression of love – this class reflects the expression of feelings of love, infatuation, and affection towards someone. This class includes texts containing such expressions.
2. Protective factors/Presence of positive social connections – this class reflects the presence of positive connections and relationships in the author’s life.
3. Protective factors/Help-seeking – this class reflects the author’s search for or request for help.
4. Protective factors/Desire for love and relationships – this class reflects the author’s desire, need, search, or aspiration for love and warm relationships.
5. Protective factors/Positive self-esteem – this class reflects positive self-perception and high self-esteem of the author. It is the opposite of the "Negative self-perception" class from the presuicidal map.
6. Protective factors/Positive beliefs – this class reflects positive attitudes and beliefs of the author. It is the opposite of the "Negative beliefs" class from the presuicidal map.
7. Protective factors/Pursuit of happiness and joy – this class reflects the desire for, search for, and aspiration to happiness and joy.
8. Protective factors/Material desires – this class reflects material needs and desires.
9. Protective factors/Presence of goals, plans for the future, favorite activity – this class reflects the presence of goals, plans for the future, favorite activities, and is associated with some kind of activity and/or actions.
10. Protective factors/Expression of happiness, joy, satisfaction – this class reflects feelings of happiness, joy, and satisfaction.
11. Protective factors/Positive dynamics – this class reflects positive dynamics and progress in the author’s emotional and/or physical state.
12. Protective factors/Supportive states and situations – this class reflects supportive states, rituals, actions, and skills of the author.

## *7. Conclusion*

In this work we show the developed methodology of creating the large-scale dataset for detecting presuicidal and anti-suicidal signals in social media texts. By using this methodology, we collect the dataset of more than 50 thousand examples. Our experiments, despite being basic, show a promising performance level of classification models. We also find typical problems raised during the annotation and model training.

In future work we plan to deal with the annotation problem of interfering irrelevant class with relevant ones. We also want to investigate the subjectivity feature of the dataset because it significantly influences the annotation process, leading to confused annotation. Also, we would like to explore the power of LLM for classification and annotation purposes. Furthermore, the augmentation strategies can be explored to give the model more examples of the 3rd rule. The same goes for overlapping lexica.

## *Информация об авторах / Information about authors*

Игорь Олегович БУЯНОВ – аспирант ФИЦ ИУ РАН, старший разработчик в MTS AI. Сфера научных интересов: обработка естественного языка, анализ пространств эмбеддингов, вычислительная психология.

Igor Olegovich BUYANOV – post graduate student at FRC CSC RAS, senior developer at MTS AI. Research interests: natural language processing, embedding space analysis, computational psychology.

Дарья Валентиновна ЯСЬКОВА – магистр психологии ННГУ им. Н.И. Лобачевского с 2018 года, старший разработчик в МТС ИИ с 2019 года. Сфера научных интересов: обработка естественного языка, распознавание именованных сущностей в специфичных доменах, методы аугментаций для текстовых данных.

Darya Valentinovna YASKOVA – master of psychology in N. I. Lobachevsky State University of Nizhny Novgorod from 2018, senior developer at MTS AI since 2019. Research interests: natural language processing, named entity recognition in specific domains, text augmentation methods.

Данил Сергеевич СЕРЕНКО является студентом кафедры математического моделирования и искусственного интеллекта РУДН имени Патриса Лумумбы, научным сотрудником Федерального исследовательского центра "Информатика и управление" Российской академии наук (ФИЦ ИУ РАН). Область научных интересов – искусственный интеллект, информационный поиск.

Danil Sergeevich SERENKO is a student at the Department of Mathematical Modeling and Artificial Intelligence of the Patrice Lumumba RUDN University, a researcher at Federal Research Center "Computer Science and Control" of the Russian Academy of Sciences. His research interests include AI, information retrieval.

Данил Николаевич ШКЕРЕДА – студент Российского государственного университета нефти и газа (национальный исследовательский университет) имени И. М. Губкина, научный сотрудником Федерального исследовательского центра "Информатика и управление" Российской академии наук (ФИЦ ИУ РАН). Сфера научных интересов: эффективное обучение больших языковых моделей, семантический анализ текстов.

Danil Nikolaevich SHKEREDA is a student at the Department of Mathematical Modeling and Artificial Intelligence of the Patrice Lumumba RUDN University, a researcher at Federal Research Center "Computer Science and Control" of the Russian Academy of Sciences. Research interests: effective training of large language models, semantic analysis of texts.

Андрей Дмитриевич ЯСЬКОВ – магистр информационных систем и технологий НГТУ им. Р. Е. Алексеева с 2017 года, разработчик в Яндекс с 2022 года. Сфера профессиональных интересов: разработка высоконагруженных веб-приложений, архитектура информационных систем, разработка интерактивных редакторов диаграмм, векторная графика, доступность веб-приложений.

Andrey Dmitrievich YASKOV – master of informatics in Nizhny Novgorod State Technical University n.a. R.E. Alekseev from 2017, developer at Yandex from 2022. Professional interests: high load web-application development, architecture of information systems, interactive editor of diagram development, vector graphics, web-application accessibility.

Илья Владимирович СОЧЕНКОВ – кандидат физико-математически наук, ведущий научный сотрудник ФИЦ ИУ РАН, ведущий научный сотрудник ИСП РАН, ведущий научный сотрудник ИППИ РАН. Сфера научных интересов: обработка естественного языка, методы информационного поиска, обработка больших массивов текстовой информации.

Ilia Vladimirovich SOCHENKOV – Cand. Sci. (Phys.-Math.), lead researcher at FRC CSC RAS, leading researcher at ISP RAS, leading researcher at IITP RAS. Research interests: Natural Language Processing, Information Retrieval, Big Data & Text Mining.